\def\year{2021}\relax
\documentclass[letterpaper]{article} 
\usepackage{aaai21}  
\usepackage{times}  
\usepackage{helvet} 
\usepackage{courier}  
\usepackage[hyphens]{url}  
\usepackage{graphicx} 
\usepackage{natbib}  
\usepackage{caption} 

\usepackage{subfigure}

\usepackage{amsmath,amsfonts,bm}

\def\eqref#1{equation~\ref{#1}}

\def\1{\bm{1}}

\def\rvtheta{{\mathbf{\theta}}}

\def\rvtau{{\mathbf{\tau}}}
\def\rvpi{{\mathbf{\pi}}}

\def\rvu{{\mathbf{i}}}

\def\rvp{{\mathbf{p}}}

\def\rvu{{\mathbf{u}}}

\def\rvy{{\mathbf{y}}}
\def\rvz{{\mathbf{z}}}

\DeclareMathAlphabet{\mathsfit}{\encodingdefault}{\sfdefault}{m}{sl}
\SetMathAlphabet{\mathsfit}{bold}{\encodingdefault}{\sfdefault}{bx}{n}

\usepackage{multirow}
\usepackage{booktabs}
\usepackage{comment}
\usepackage{color}

\definecolor{amaranth}{rgb}{0.9, 0.17, 0.31}
\definecolor{azure(colorwheel)}{rgb}{0.0, 0.5, 1.0}

\title{Contextual Temperature for Language Modeling}
\author {
    Pei-Hsin Wang,\textsuperscript{\rm 1}
    Sheng-Iou Hsieh,\textsuperscript{\rm 1}
    Shih-Chieh Chang,\textsuperscript{\rm 1} \\
    Yu-Ting Chen, \textsuperscript{\rm 2}
    Jia-Yu Pan, \textsuperscript{\rm 2}
    Wei Wei, \textsuperscript{\rm 2}
    Da-Chang Juan \textsuperscript{\rm 2} \\
}
\affiliations {
    \textsuperscript{\rm 1} National Tsing Hua University, Hsinchu, Taiwan 
    \textsuperscript{\rm 2} Google, Mountain View, CA, USA \\
    \{peihsin, steins1111\}@gapp.nthu.edu.tw, scchang@cs.nthu.edu.tw, \\
    \{yutingchen, jypan, wewei, dacheng\}@google.com
}
\begin{document}

\maketitle

\begin{abstract}
Temperature scaling has been widely used as an effective approach to control the smoothness of a distribution, which helps  the model performance in various tasks. Current practices to apply temperature scaling assume either a fixed, or a manually-crafted dynamically changing schedule. However, our studies indicate that the individual optimal trajectory for each class can  change with the context. To this end, we propose contextual temperature, a generalized approach that learns an optimal temperature trajectory for each vocabulary over the context. Experimental results confirm that the proposed method significantly improves state-of-the-art language models, achieving a perplexity of 55.31 and 62.89 on the test set of Penn Treebank and WikiText-2, respectively. In-depth analyses show that the behaviour of the learned temperature schedules varies dramatically by vocabulary, and that the optimal schedules help in controlling the uncertainties. These evidences further justify the need for the proposed method and its advantages over fixed temperature schedules.
\end{abstract}

\section{Introduction}
\label{sec:Introduction}
Temperature scaling is often used along with the Softmax layer in various tasks, such as knowledge distillation, model calibration, and natural language generation \cite{krizhevsky2012imagenet, bahdanau2014neural, Hinton2014KD, Guo2017Cali, Hu2017control, Caccia2018gan}. A widely adopted technique is to apply a temperature as a denominator to the logits of the Softmax layer. Specifically, given a temperature $\tau$, when $\tau \to \infty$, the distribution becomes more uniform, thus increasing the uncertainty. Contrarily, when $\tau \to 0$, the distribution collapses to a point mass. 
Although temperature scaling has been justified to achieve great success, existing methods use temperature scaling in extremely limited ways. They either assume a constant temperature throughout the training \cite{Norouzi2016raml, ma2017sqdml, chen2018da}, or use a fixed schedule to control the temperature \cite{Hu2017control}. Most importantly, none of them studies the effects on different vocabulary tokens when the temperature changes.
We propose \textit{contextual temperature}, a generalized temperature scaling method which takes both the history of a sequence and a particular word token into consideration. 
Through optimizing the use of temperature scaling by the change of contexts, contextual temperature is able to generate an unique temperature for each token.

\begin{figure*}[!h]
\begin{center}
\subfigure[]{
\label{fig:T-training}
\includegraphics[width=0.49\textwidth]{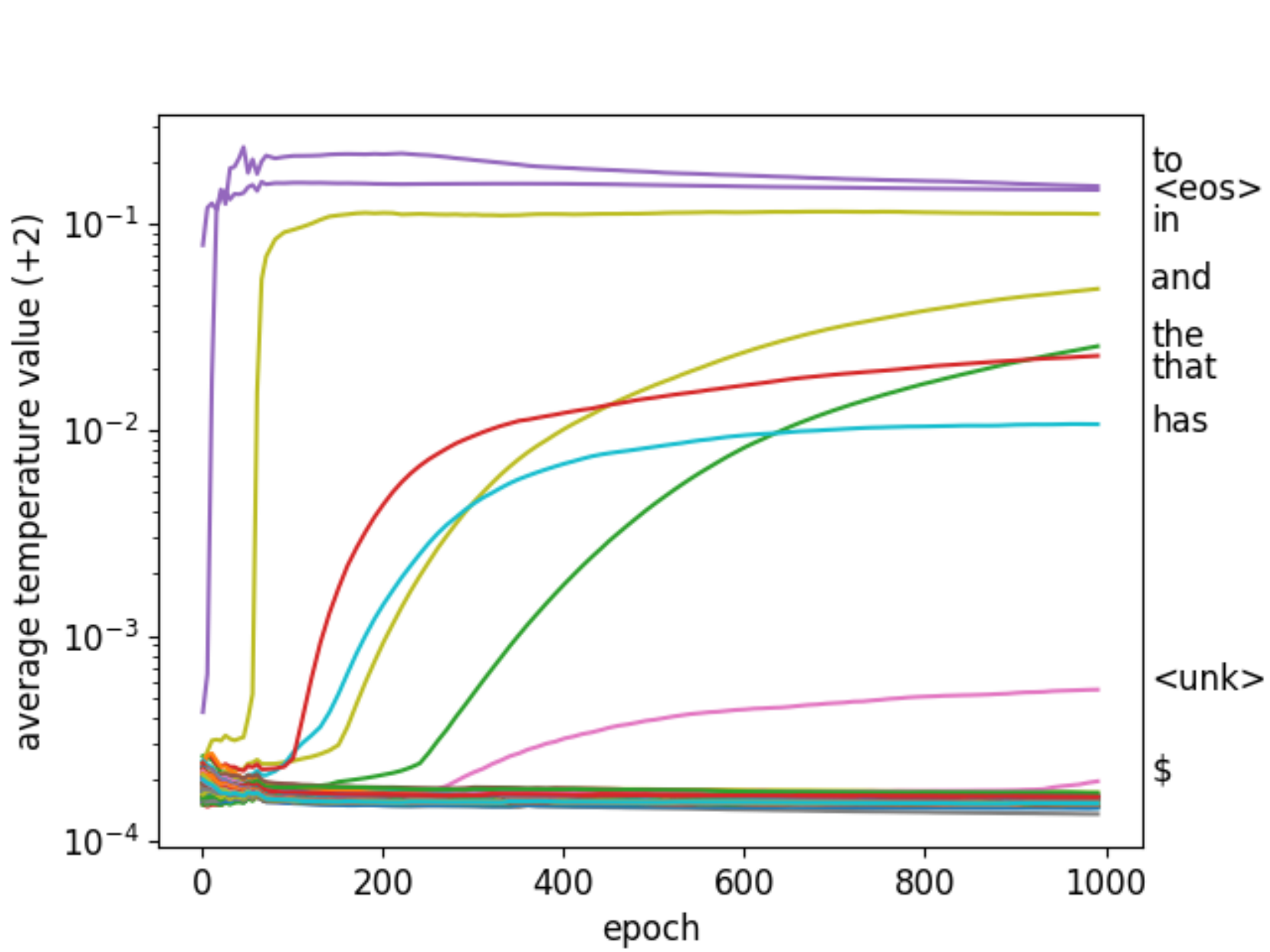}}
\subfigure[]{
\label{fig:T-pos-15}
\includegraphics[width=0.49\textwidth]{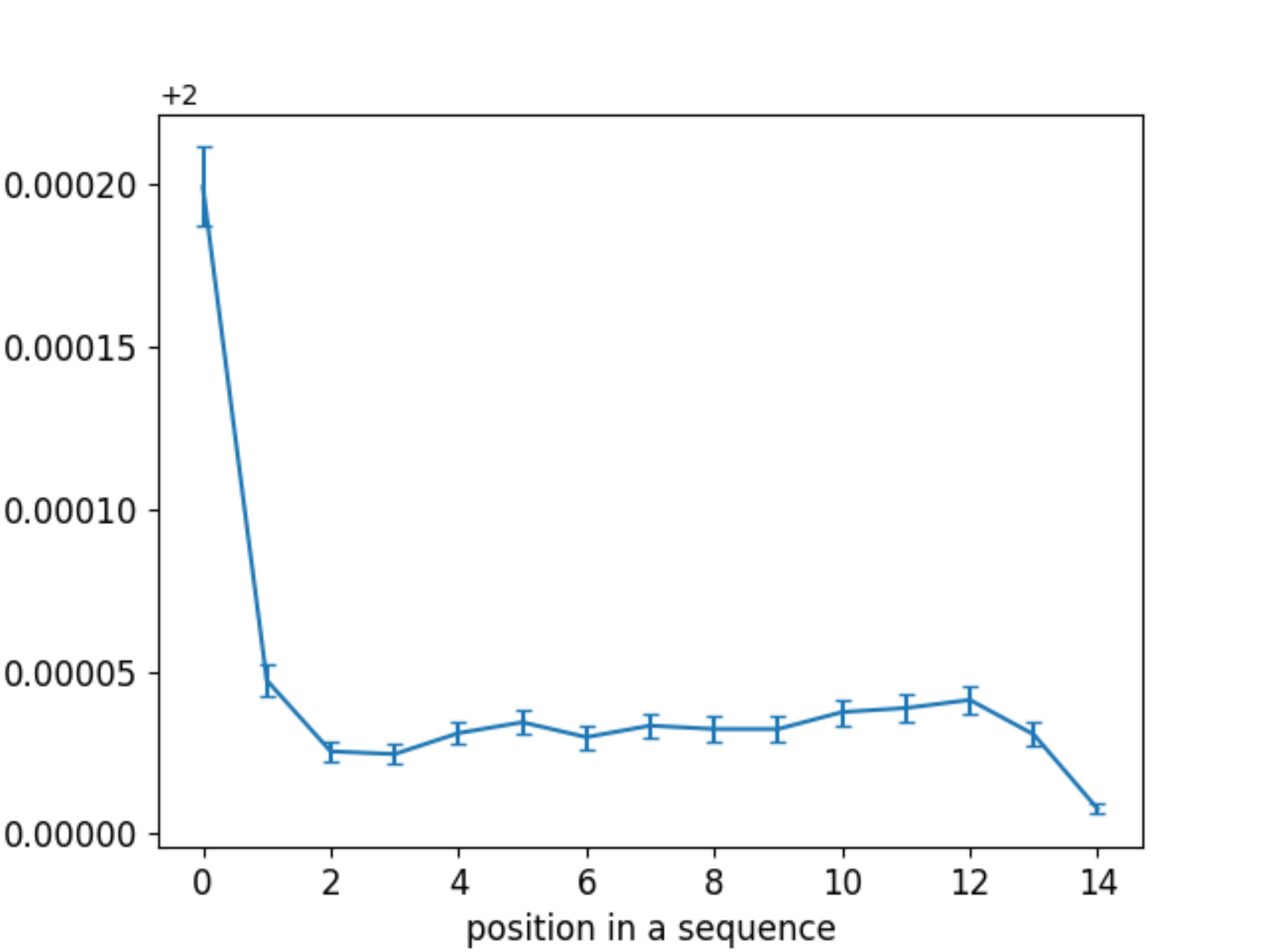}}
\caption{(a) Temperature of each word learned using the proposed method over training epochs. Each line in the figure represents a distinct token ranked by frequency. The y-axis shows the tokens' temperature values. Note that the actual value equals to the value on the y-axis plus 2; we eliminate the integer part (plus 2) for a better visualization. (b) Means and confidence intervals of the temperature vector (y-axis) over positions in a sentence (x-axis). As in this figure, the average temperature is high at the beginning of a sentence and gradually decreases towards latter positions in a sentence. 
 }
\end{center}
\end{figure*}

Figure \ref{fig:T-training} shows the optimized temperature for each token during the course of training. As shown in the figure, some certain words have a distinct heating-up temperature scaling, while the majority of words have a scaling that gradually cooling down the temperature. We argue that existing methods limit themselves to some fixed schedules, and thus have great difficulty to generalize. In addition, another example can be observed in Figure 1(b), which indicates that the average of the temperature drops, as the length of the context increases. This suggests that the temperature mechanism helps promote stochasticity in the beginning of a sentence, then gradually suppressing the uncertainty until the end. All of these suggest the use of a more generalized temperature mechanism with the advantage of being able to deal with these phenomena.

Experiments on language modeling exhibit significantly improved performances on both Penn Treebank and WikiText-2 datasets. Consistent improvements are shown on both validation and testing splits. In addition, we conduct comprehensive analyses and ablation studies to confirm the improvements. We observe that the proposed method is capable of controlling the uncertainties as the patterns of contexts change, allowing language models to achieve much better performances. To the best of our knowledge, this work is the first systematic work that studies the role of temperature changing over context for training language models on a per token basis. The experiment results suggest a new way of training sequential models with discrete outputs, that is, using parameterized temperatures.  We have also established the link between the control of model uncertainty and the use of temperatures, paying the way for extensions on tasks that require such long-term control. 

\section{Related Work}
\label{sec:RelatedWork}
\subsection{Language Modeling}

Language modeling aims at predicting the next word $x_t$ in a sentence given a sequence of history $x_{1:t-1}=x_1,...,x_{t-1}$. 
A language model factorizes $P(x_{1:t})$, the joint probability of $x_{1:t}$, as the product of a series of conditional probabilities, 
as specified below:
\begin{equation}
\label{eq:lm}
P(x_{1:t})=\prod_{i=1}^{t}P(x_i|x_{1:i-1})
\end{equation}

Modern neural language models are comprised of two parts: a mapping function and a probability function over tokens. A mapping function $\rvtheta^{emb} \in \mathbb{R}^{|V|\times d}$ maps a token $x_t$ into a real-value vector $x_{t}^{emb} \in \mathbb{R}^d$, where $d$ is the dimension of the vector and $|V|$ is the size of vocabulary. A probability function converts a $d$ dimensional vector into a $|V|$ dimensional vector using a weight matrix $\rvtheta^{model} \in \mathbb{R}^{d\times |V|}$. Afterwards, a Softmax function $\sigma$ is used to obtain a probability distribution over tokens. That is, at timestep $t$, we have
\begin{equation}
\label{eq:prob}
P(x_{t}|x_{1:t-1}) = \sigma(f(x_{1:t-1} ; \rvtheta^{emb} \cup \rvtheta^{model}))
\end{equation}
where $f$ is a nonlinear mapping parameterized by $\rvtheta^{emb}$ and $\rvtheta^{model}$. Current language models adopt various architectures as $\rvtheta^{model}$, including the earlier feed-forward networks~\cite{bengio2003neural}, recent recurrent-based ones~\cite{hochreiter1997lstm, mikolov2010rnn, zilly2016rnnhighway}, convolutional-based ones~\cite{dauphin2017gatedcnn}, and attention-based ones \cite{Dai2019TransXL}.

\subsection{Softmax Layer}
A Softmax layer $\sigma$ normalizes a $|V|$ dimension, real-value vector $\rvz$ to make it sum to $1$. Here, $\rvz=f(x_{1:t-1};\rvtheta^{emb} \cup \rvtheta^{model})$ following Equation \ref{eq:prob}, and $z_i$ is the $i$-th element in $\rvz$.
\begin{equation}
\label{eq:softmax}
\sigma (\rvz)_i = \frac{z_i}{\sum_j^K e^{z_j}} 
\end{equation}

Recent progress suggests that Mixture of Softmaxes (MoS)~\cite{yang2018Mos} significantly improves the performance by first computing multiple Softmax distributions, and summing them up through a set of weights to provide the final probability distribution. Specifically, a set of $M$ matrices $W_m$ is applied to $\rvz$, yielding $\rvz_{m}=\rvz^T \cdot W_m$, where $W_m \in \mathbb{R}^{d \times |V|}$. 
On the other hand, the weights of $M$ Softmaxes, denoted $\pi$, are parameterized by $W_p \in \mathbb{R}^{h\times M}$, $h$ the dimension of RNN output.
The probability distribution under the MoS model is thus
\begin{equation}
\label{eq:mos}
\small
   P_{MoS}(x_t|x_{1:t-1} ; \Theta) = \sum_m^M \rvpi_{m} \cdot \sigma (\rvz_{m})
\end{equation}
where $\Theta = \cup_{m=1}^M W_m \cup \rvtheta^{emb} \cup \rvtheta^{model} $.

\subsection{Temperature Scaling}
Temperature, denoted $\tau$, often serves as a hyper-parameter in the Softmax layer to control the smoothness of the distribution. Applying temperature scaling on $\rvz$ gives 
\begin{equation}
\label{eq:def_yt}
\small
   P(x_t|x_{1:t-1}; \rvtheta^{emb}, \rvtheta^{model}, \tau) = \sigma (\rvz/{\tau})
\end{equation}
Here $\tau$ is a vector with dimension $|V|$ same as $\rvz$, and is normally $\tau$ is set to $1$. 

Various usages of temperature scaling have been explored in recent progresses, breaking the limit of treating temperature as a hyper-parameter. 
Below we divide related works into three categories: (a) constant temperature, where each element in $\tau$ has the same value, so each element $z_i$ in $\rvz$ is divided by the same temperature, (b) dynamic temperature over training iterations, where $\tau$ is constant in one single iteration, but has different value in every iteration, and (c) dynamic temperature over word position, where $\tau$ is dynamic in one iteration, i.e., each element $z_i$ in $\rvz$ has its own temperature value, and besides, $\tau$ could change in every iteration.

\paragraph{Constant Temperature.} Earlier works can be traced back to model distillation \cite{Hinton2014KD}, where $\tau$ is a hyper-parameter and chosen empirically.
Constant temperature is also used during training \cite{Norouzi2016raml, ma2017sqdml} to control the degree of augmentation. 
Other works incorporating $\tau$ at inference time include model confidence calibration \cite{Guo2017Cali}, and controlling trade-off between quality and diversity in text generation tasks \cite{Caccia2018gan}.

\paragraph{Dynamic Temperature Over Training Iterations.} Most works adopt dynamic temperature through a manually-crafted schedule. 
Notably, \citet{Hu2017control} use an approximation based on Softmax with a decreasing temperature to enable gradient propagation. 
Similar techniques are adopted in gumbel-softmax \cite{Jang2017gumbel} to allow gradients to pass through discrete sampling objectives. 
In addition, studies show that with a heating up temperature scaling, embedding vectors are more compact \cite{zhang2018heated}.

\paragraph{Dynamic Temperature Over Word Position.}
Another work that is closely related to our work, is the adaptive temperature over an attention model \cite{Lin2018sact}. At each timestep, the model learns to output a dynamic temperature to control the softness of the attention, based on the information of decoding at the current step as well as the attention in previous steps. We note that contextual temperature further learns the temperature for each vocabulary in the output distribution.

\section{Methods}
\label{sec:Methods}
\subsection{Contextual Temperature}
Contextual temperature is a mechanism that chooses the optimal temperature by considering the ``context'' of a word $x_t$. A context of a word includes not only the history $x_{1:t-1}$, but also the specific vocabulary $k$ that we calculate the probability on. Such a mechanism allows us to parameterize the temperature vector $\rvtau$ using a deep neural network and adapt the softness of the Softmax layer.

Our temperature vector $\rvtau \in \mathbb{R}^{|V|}$ is generated from the non-linear mapping function $f$ as described in Section \ref{sec:RelatedWork}. 
Although $f$ can be any sequential models, we parameterize it by AWD-LSTM~\cite{Merity2018AWD}. 
We then multiplying the output of $f$ by $W_{\tau}$ 
Please note that using a single matrix may increase the number of parameters significantly, and thus in practice one can choose to factorize it into two matrices. Finally, we scale the temperature vector using a Softmax layer over the dimension of vocabulary, bounding its range in [$\frac{\alpha}{\beta}$, $\frac{1+\alpha}{\beta}$].
\begin{equation}
\label{eq:def_Tnew}
    \rvtau = \frac{\sigma(f(x_{1:t-1}; \rvtheta^{emb}, \rvtheta^{model})^T \cdot  W_{\tau})+\alpha}{\beta}
\end{equation}

\subsection{Contextual Temperature MoS}
\label{subsec:CT-MoS}
We use the Contextual Temperature Mixture of Softmaxes architecture (CT-MoS) for language modeling. The CT-MoS model extends Equation~\ref{eq:mos} by adding contextual temperature in Equation~\ref{eq:def_Tnew}.
\begin{equation}
\label{eq:CT-MoS}
\small
   P_{CT-MoS}(x_t = k|x_{1:t-1} ;\Theta) = 
   \sum_m^M \rvpi_{m} \cdot  
   \sigma (\rvz_{m} \oslash \rvtau))
\end{equation}
where $\oslash$ represents element-wise division, $\Theta=\cup_{m=1}^M W_m \cup \rvtheta^{emb} \cup \rvtheta^{model} \cup W_{\tau_1} \cup W_{\tau_2}$ denotes the total parameters of CT-MoS, and $W_{\tau} = W_{\tau_1} \cup W_{\tau_2}$. 

\begin{figure}[!ht]
\begin{center}
\includegraphics[width=0.45\textwidth]{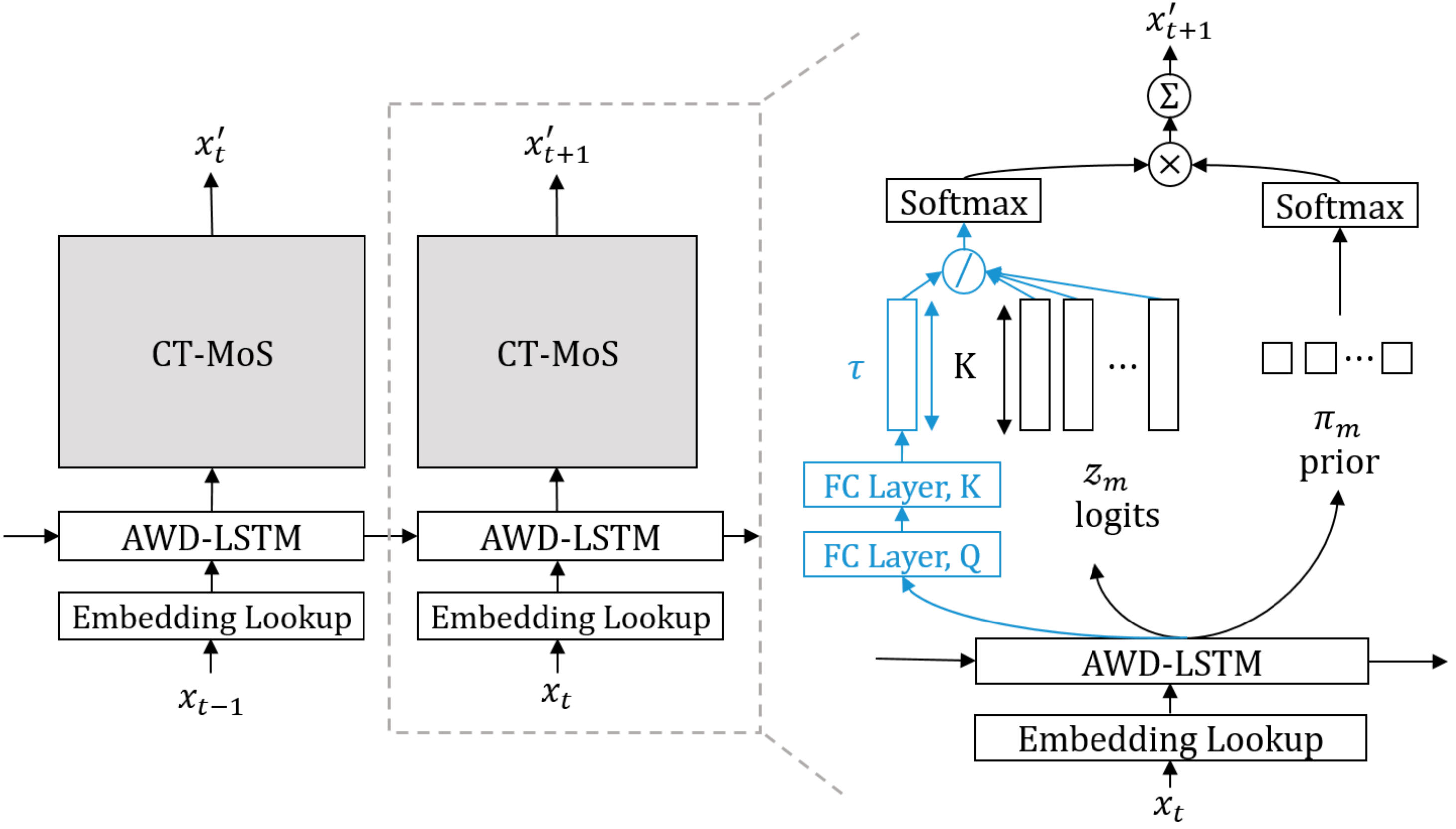}
\caption{The architecture of the proposed CT-MoS model. Black components are those the same as the MoS model, while the blue ones are the newly added ones in the proposed approach.}
\label{fig:archi}
\end{center}
\end{figure}

Compared to prior works, the proposed method is not only dynamic over training iterations, but also dynamic over word positions 
at current timestep. Most importantly, for the same token but at different position of context, the proposed method learns a different temperature vector dependent on its history context. The detailed architecture is illustrated in Figure~\ref{fig:archi}, which highlights the difference between the proposed CT-MoS model and the MoS model. 

\begin{figure*}[!h]
\begin{center}
\subfigure[$i=1$]{
\label{fig:grad-z1}
\includegraphics[width=0.28\textwidth]{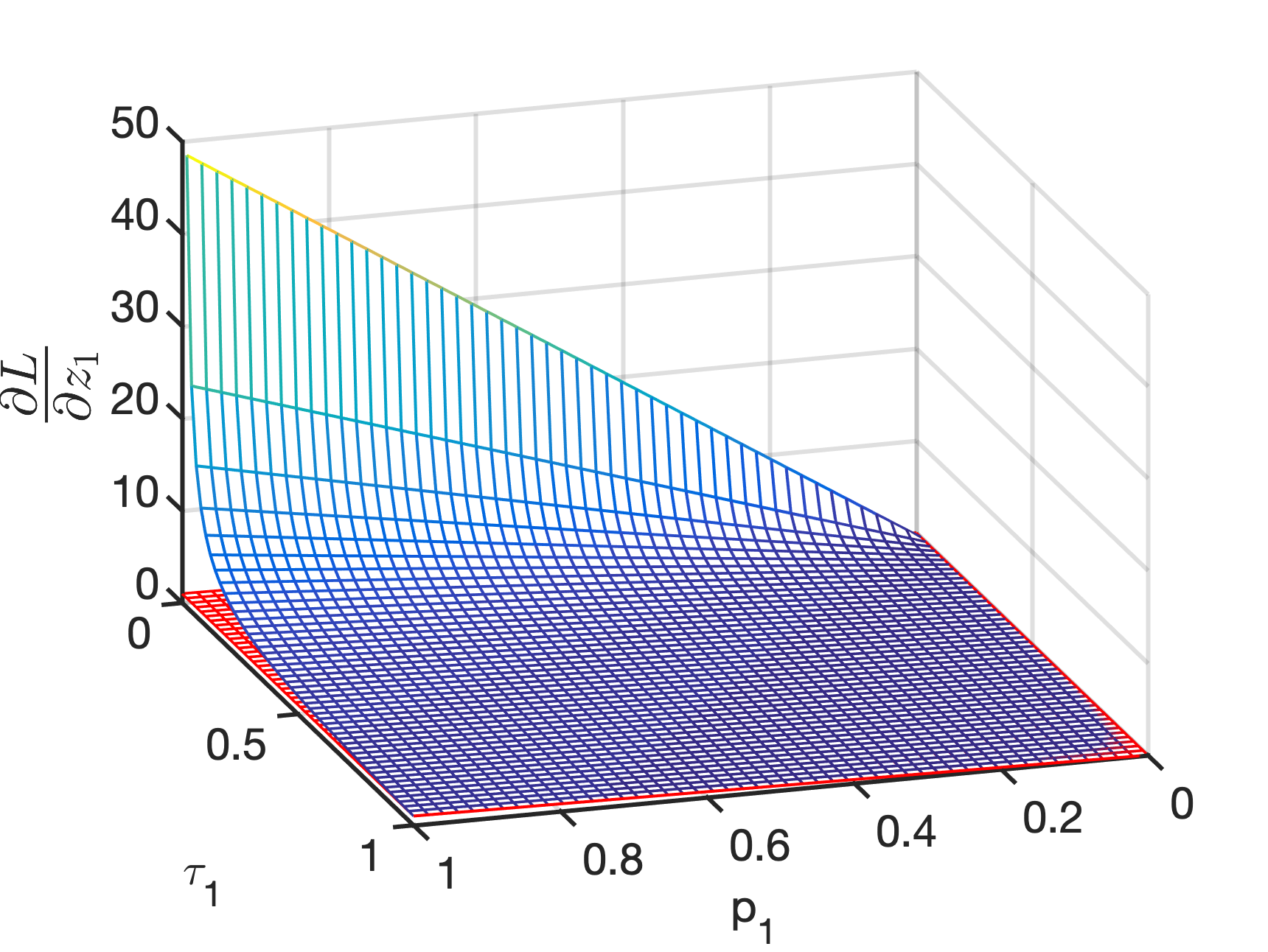}}
\subfigure[$i=0$]{
\label{fig:grad-z0}
\includegraphics[width=0.28\textwidth]{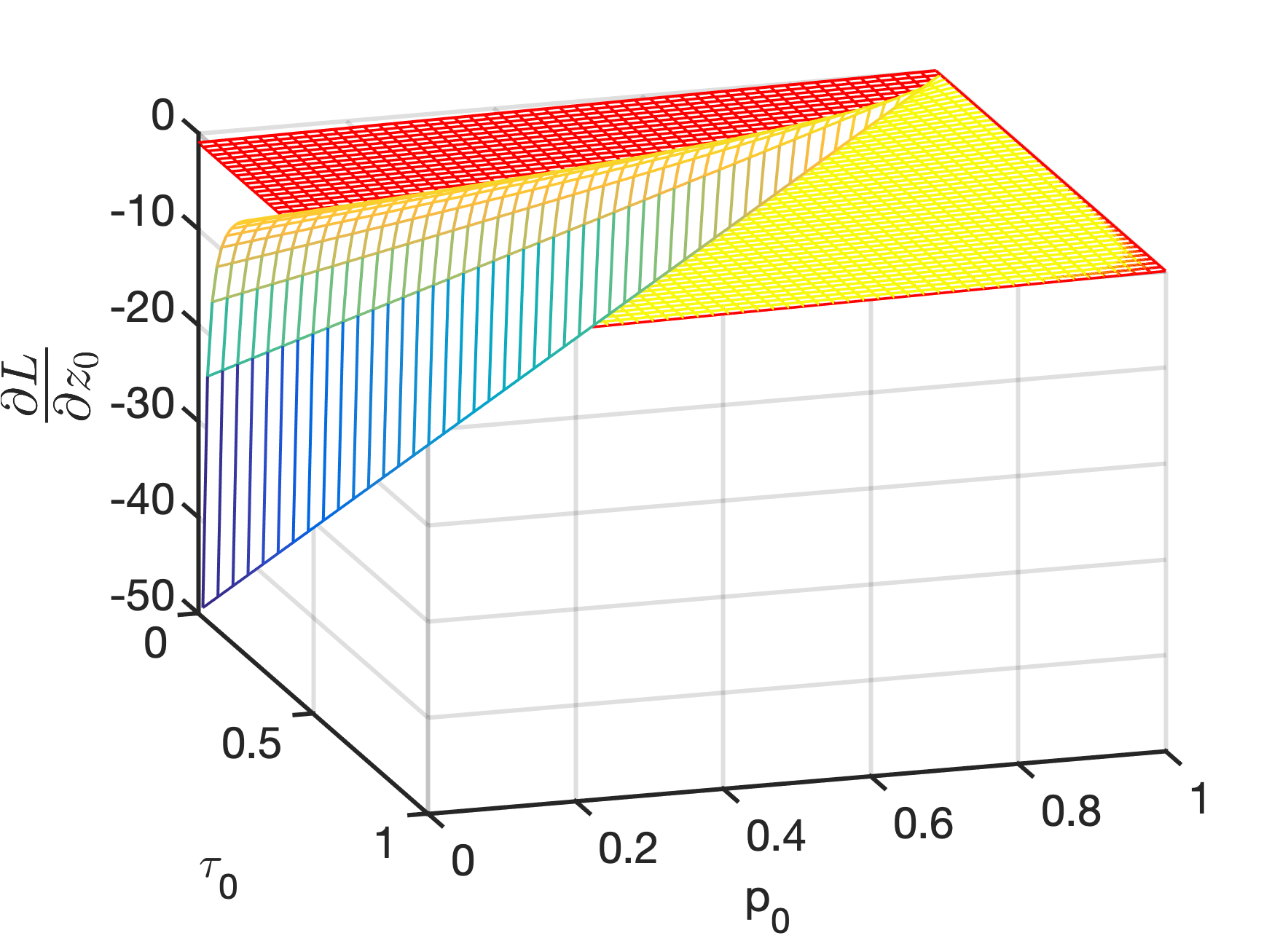}
}
\caption{
Gradients of loss with respect to (a) logit $z_0$ and (b) logit $z_1$.
In each figure, the x-axis is the probability $p_i$ of class $i$, y-axis is the temperature value $\tau_i$ of class $i$, and z-axis is the gradient $\frac{\partial L}{\partial z_i}$. The colorful mesh represents the gradients when the contextual temperature is applied, while the red mesh represents the case without temperature. 
}
\label{fig:grad-logits}
\end{center}
\end{figure*}

\begin{figure*}[!h]
\begin{center}
\subfigure[$z_0 \in \mathbb{R}^-$, $i=0$]{
\label{fig:grad-t0-1}
\includegraphics[width=0.23\textwidth]{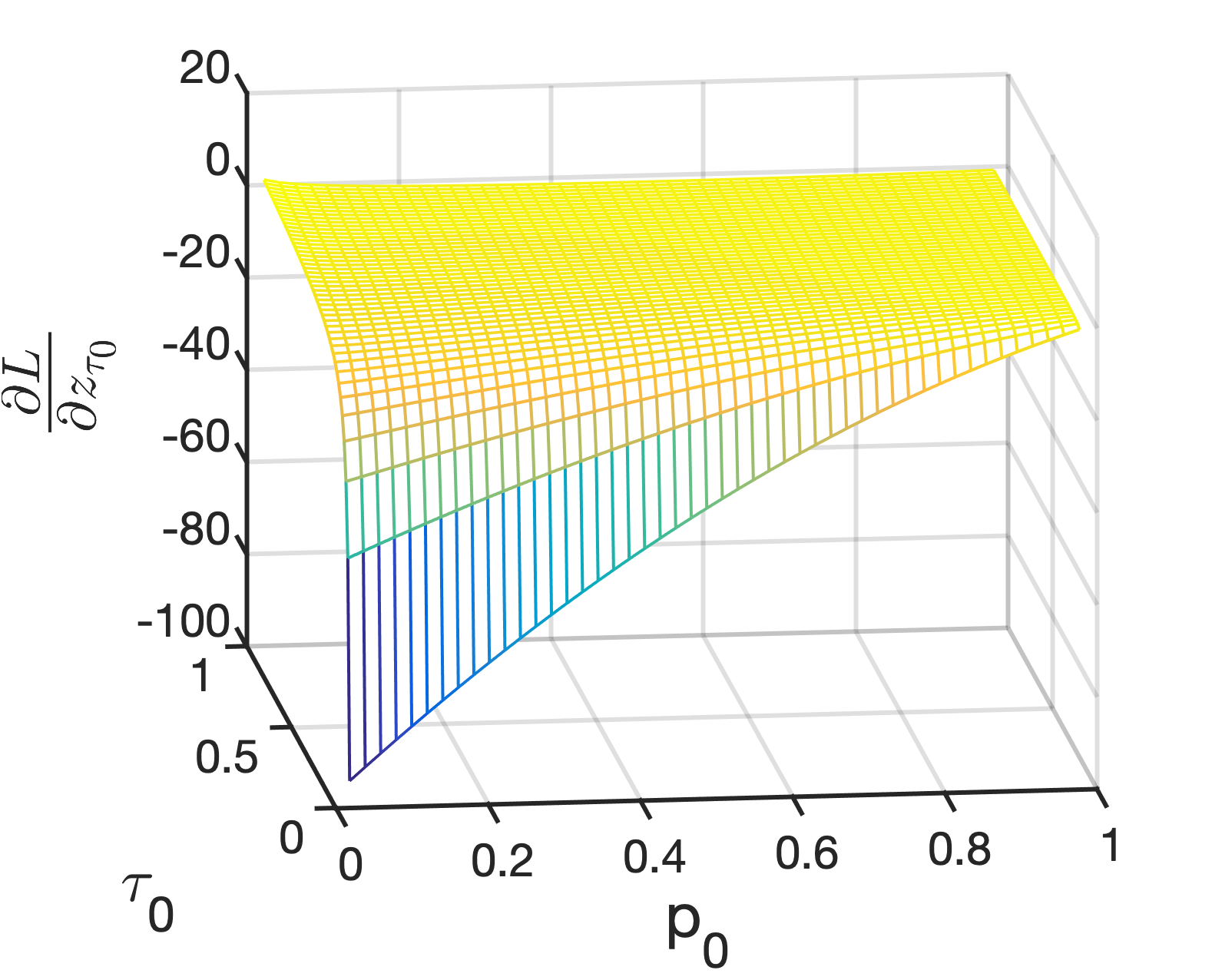}}
\subfigure[$z_0 \in \mathbb{R}^-$, $i=1$]{
\label{fig:grad-t1-1}
\includegraphics[width=0.23\textwidth]{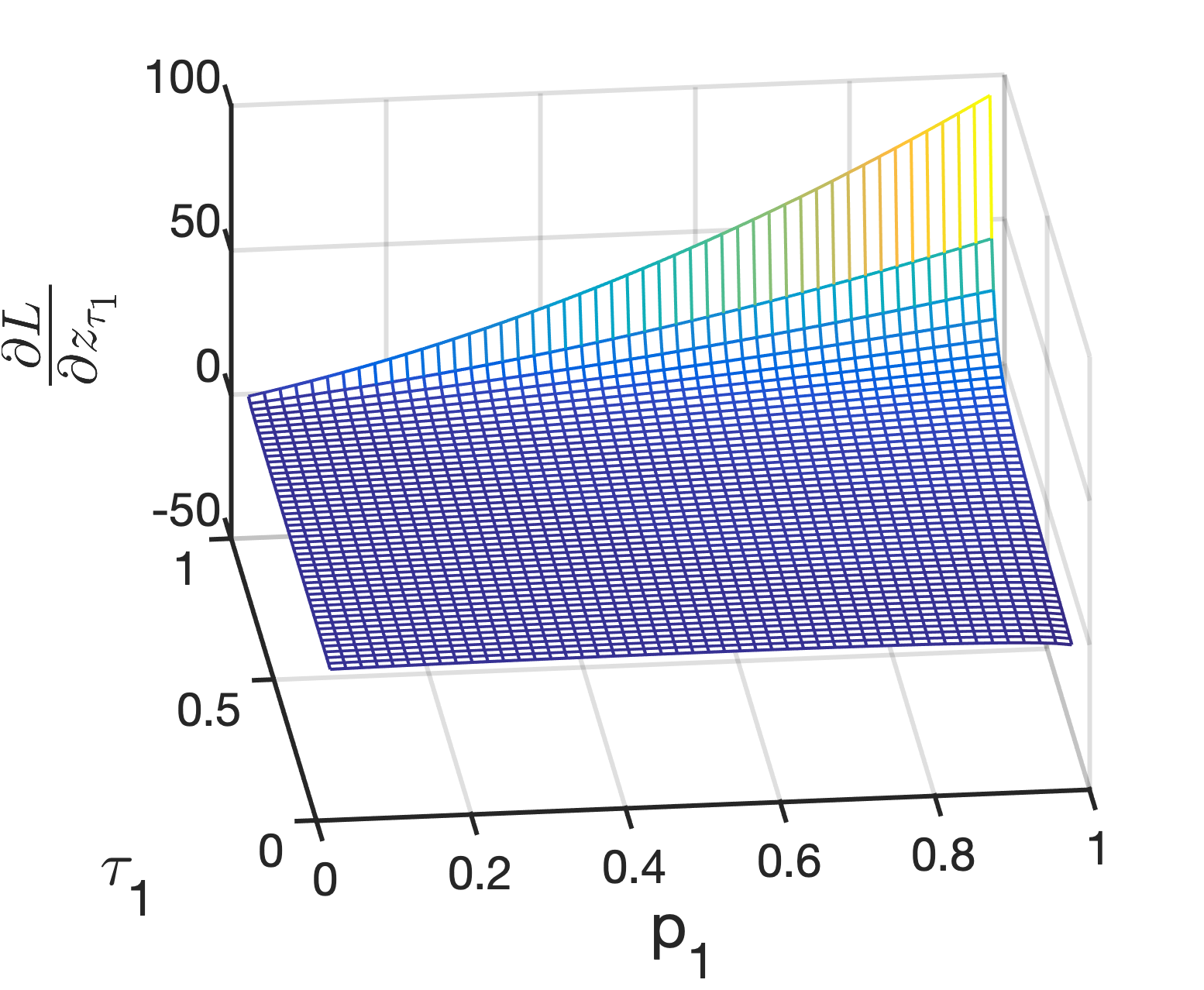}}
\subfigure[$z_0 \in \mathbb{R}^+$, $i=0$]{
\label{fig:grad-t0-4}
\includegraphics[width=0.23\textwidth]{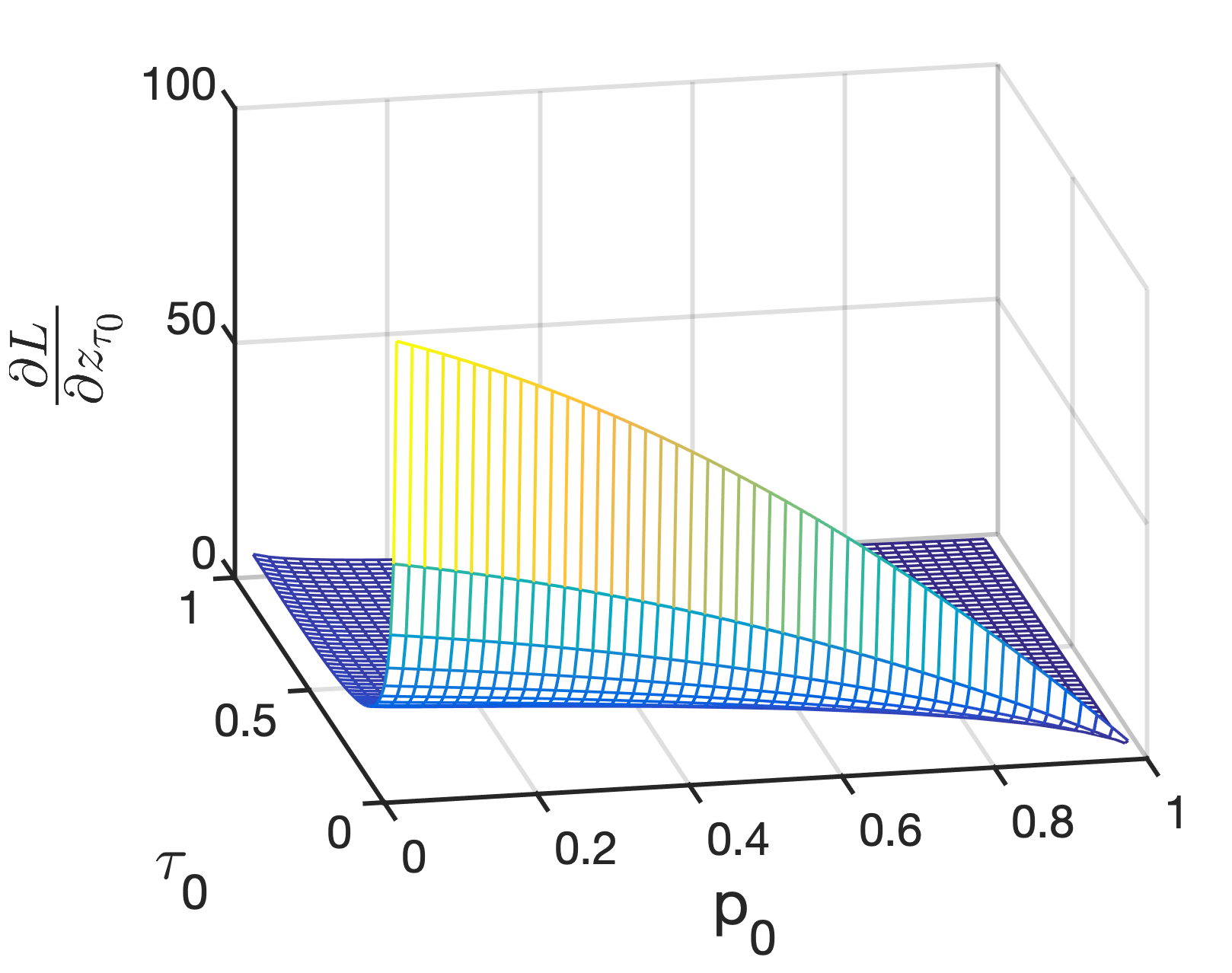}}
\subfigure[$z_0 \in \mathbb{R}^+$, $i=1$]{
\label{fig:grad-t1-4}
\includegraphics[width=0.23\textwidth]{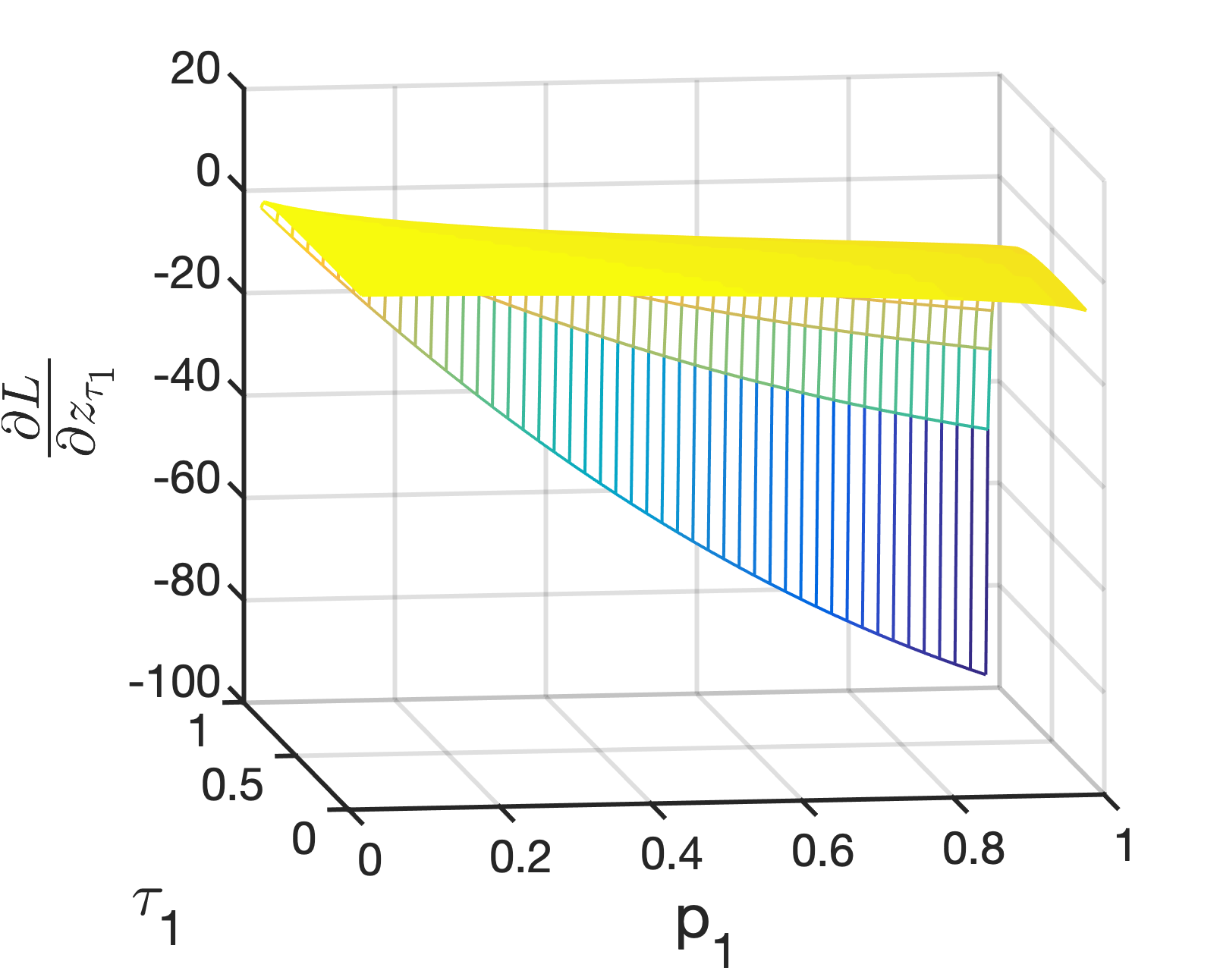}}

\caption{
Gradients of loss with respect to $z_{\tau_i}$. The x-axis is the probability $p_i$ of class $i$, y-axis is the temperature value $\tau_i$ of class $i$, and z-axis is the gradient $\frac{\partial L}{\partial z_{\tau_i}}$.
}
\label{fig:grad-taus}
\end{center}
\end{figure*}

We adopt the same regularization techniques in MoS and AWD-LSTM. Our loss function thus consists of four terms: Cross Entropy ($\mathcal{H}$), Activation Regularization (AR), Temporal Activation Regularization (TAR), and Weight Decay (WD). AR penalizes high values of outputs, TAR is used to prevent outputs from changing drastically between time steps, and WD prevents the model from overfitting. Here $\gamma1$, $\gamma2$ and $\gamma3$ are constants to scale regularization terms, and $m$ is the dropout mask.

\begin{equation}
\small
\begin{split}
     \mathcal{L} (\Theta) & = \underbrace{\frac{\sum^{K}_i \tau_i}{K}}_{\text{LS}} \mathcal{H}(\hat{\rvy},\rvy) + \gamma_1 \underbrace{ L_2(m \odot f(x_{1:t-1}; \rvtheta))}_{\text{AR}} \\
    & + \gamma_2 \underbrace{ L_2(f(x_{1:t-1}; \rvtheta) - f(x_{1:t}; \rvtheta))}_{\text{TAR}} + \gamma_3 \underbrace{ L_2(\Theta)}_{\text{WD}}
\end{split}
\label{eq:Regularization}
\end{equation}

The uniqueness in our setting is the Loss Scaling term (LS).
Using temperature scaling makes gradients of $\mathcal{H}$ disproportional to that of the other three terms, since the three terms do regularization on parameters before temperature scaling. This difference leads to the unbalance between four objectives. Therefore, we scale $\mathcal{H}$ in order to redress the balance. A similar phenomenon is reported by~\citeauthor{Hinton2014KD}. We empirically find that scaling $\mathcal{H}$ with the average of the temperature vector works well in our setting.

\subsection{Effects of Contextual Temperature}
In this section, we discuss how the proposed method effects the logits $\rvz$ and the temperature itself $\rvtau$, through illustrating their corresponding gradients. We consider a language modeling task with a small vocabulary of only two words, i.e., $|V|=2$. In this setting, the dimensionality of logits $\rvz$ is $2$ and so is that of the temperature vector $\rvtau$. The range of $\rvtau$ is set to $[0,1]$, that is, $(\alpha, \beta)=(0,1)$. 

\paragraph{Gradient of the logit.}
At a given input, assume that the ground-truth token is $i=0$, thus the cross-entropy loss is $L = -\ln p_0$, where $p_0$ is the model's output probability for word token $i=0$.
In this case, the gradients of logits $\rvz$ are illustrated in Figure~\ref{fig:grad-logits} (the derivation of the gradients is in Appendix). 
In Figure~\ref{fig:grad-z1}, we consider the gradient of $z_1$ of token $i=1$. When no temperature mechanism is applied, it can be seen that the magnitude of the gradient is bounded within [0, 1] (as shown by the red mesh), with the largest magnitude $1$ (most aggressive update) happens when the probability $p_1$ is closer to $1$ (note that the ideal value for $p_1$ should be close to 0, as the other token $i=0$ is assumed to be the ground truth).

On the other hand, when the contextual temperature is applied, the gradient can now be set to a much more substantial magnitude by setting a smaller temperature value $\tau_1$ (as shown by the blue mesh).  The additional flexibility of magnifying the gradient by temperature enables the model to be more efficient (or aggressive) in adjusting the parameters to reduce the training loss.

\paragraph{Gradient of the temperature.}
Next, we analyze how the model with contextual temperature updates the value of the temperature.  We recall that Equation~\ref{eq:def_Tnew} defines the formulation of the temperature.  We leave the detailed derivation of the gradient of the temperature in Appendix.
Figure~\ref{fig:grad-taus} visualizes the gradients $\partial L / \partial z_{\tau_0}$ and $\partial L / \partial z_{\tau_1}$ when $z_0$ is either positive or negative.

We will use the case in Figure~\ref{fig:grad-t0-1} (where $z_0 \in \mathbb{R}^-$) as an example to explain the effect of the gradient on the temperature.  In this case, if $p_0$ is close to $0$, then we expect that the model should be more aggressive to update the parameters (since we assume that the ground-truth class is $i=0$, meaning that $p_0$ should ideally be close to $1$).  This aggressive response is indeed visible in the figure, showing values of larger magnitude when $p_0 \to 0$.

The amount of update on the temperature $\tau_0$ also depends on its current value.  In the same Figure~\ref{fig:grad-t0-1}, let's consider a fixed $p_0$, say, $p_0 = 0.1$.  In this case, as described above, the model wants to increase the value of $p_0$ to closer to $1$.  To do that, the model will attempt to increase the term $e^{z_0 / \tau_0}$.  Since $\tau_0$ is a positive value in [0, 1] and $z_0 < 0$ in this case, to maximize $z_0 / \tau_0$, the optimal is to have the temperature value $\tau_0 \to 1$.  When the model is updating the value of $\tau_0$, if its current value is already close to $1$, then the gradient will be small, since it is already close to the optimal value.  On the other hand, if the temperature $\tau_0$ is still far from $1$, then the gradient will update it's value more aggressively.  This behavior is exactly visualized in the figure.

\section{Experiments}
\label{sec:Experiments}
\subsection{Experimental Setups}
We evaluate the proposed method on two widely-used datasets for language modeling: Penn Treebank (PTB) \cite{Marcus1993PTB, mikolov2011lmtech} and WikiText-2 (WT2) \cite{Merity2017WT2}. 
The PTB dataset contains 929k training, 73k validation and 82k test tokens. The vocabulary size is capped at 10,000 most frequent unique tokens, while the rest of tokens are replaced with the $<$unk$>$ token. We follow common practices to pre-process the dataset \cite{mikolov2011lmtech}: (a) words are lower-cased, (b) numbers are replaced with ``N'', (c) newlines are replaced with $<$eos$>$ and (d) punctuation is removed. 
WikiText-2 is derived from Wikipedia articles and released as a popular option to train language models. WT2 contains 2M training tokens and a vocabulary of around 33k tokens. Compared to PTB, WT2 is roughly two times larger in sample size and three times larger in vocabulary size. 

We conduct experiments on PTB and WT2 using one and four 1080 Ti GPUs, respectively. The environment we use is PyTorch~\cite{paszke2017automatic}. We follow the training configurations as reported in the MoS paper and their github\footnote{\url{https://github.com/zihangdai/mos}}.
For both PTB and WT2, we use the same number of parameters as MoS. That is, we use three layers of LSTM with embedding sizes of 960-960-620 on PTB; and for WT2, we use three layers of LSTM with embedding sizes of 1150-1150-650.
The regularization terms, $\gamma_1, \gamma_2,$ and $\gamma_3$ are 2.0, 1.0 and 1.2$e^{-6}$, respectively. The number of Softmaxes to be mixed is $M=15$. 
Furthermore, we perform normalization described in Equation \ref{eq:def_Tnew} to ensure each element in $\tau$ is positive.
We have tried several different values for $\alpha$ and $\beta$, and find that $(\alpha, \beta) = (1, 0.5)$ work best in all experiments.

\subsection{Results}
We first show the results on the Penn Treebank dataset in Table~\ref{LM-PTB}. We compare the proposed method with MoS and AWD-LSTM, and the performance  is compared both with and without fine-tuning and dynamic evaluation \cite{Krause2017Dynamic}. Since the original MoS model has approximately 22M parameters, to make a fair comparison, we augment its number of parameters to have 24M parameters. We increase the size of each layer proportionally, giving the word embedding size $d=300$, and making the sizes of the three LSTM layers be $1030,1030$ and $670$. We denote this augmented model as MoS$+$. We show that our CT-MoS model outperforms AWD-LSTM, MoS and MoS$+$ models on both validation and test sets. The conclusion holds for the default setting with fine-tuning, the one without fine-tuning, and the one with dynamic evaluation.  Our best model achieved 48.12 perplexity on the validation set and 47.42 on the test set, beating the best state-of-the-art model of MoS with a great margin.

We also present Table~\ref{LM-WT2} to provide the results on the WikiText-2 dataset. Here we see a similar pattern with the PTB dataset that CT-MoS outperforms the state-of-the-art models. Note that when using dynamic evaluation, our model achieves comparable improvements (32\%) to MoS (33\% improvement) and AWD-LSTM (32\% improvement).

\begin{table}[!ht]
\small
\begin{center}
\begin{tabular}{l|cccl}
\toprule
\multicolumn{1}{c|}{Model}  & \#Param & Validation    & Test   \\
\midrule
AWD-LSTM w/o finetune   & 24M   & 60.7  & 58.8  \\
AWD-LSTM                & 24M   & 60.0  & 57.3  \\
AWD-LSTM $^\dagger$     & 24M   & 51.6  & 51.1  \\
\midrule
MoS w/o finetune        & 22M   & 58.08 & 55.97  \\
MoS                     & 22M   & 56.54 & 54.44  \\
MoS $^\dagger$          & 22M   & 48.33 & 47.69  \\
\midrule
MoS$+$ w/o finetune     & 24M   & 59.72 & 57.43  \\
MoS$+$                  & 24M   & 58.54 & 56.36  \\
MoS$+ ^\dagger$         & 24M   & 50.49 & 49.81  \\
\midrule
CT-MoS w/o finetune & 24M   & \bf56.95  & \bf54.69 \\
CT-MoS              & 24M   & \bf55.31  & \bf53.2  \\
CT-MoS$^\dagger$    & 24M   & \bf48.12  & \bf47.42 \\
\bottomrule
\end{tabular}
\end{center}
\caption{Performance comparison on the Penn Treebank (PTB) dataset. $\dagger$ indicates using dynamic evaluation.}
\label{LM-PTB}
\end{table}

\begin{table}[!ht]
\small
\begin{center}
\begin{tabular}{l|cccl}
\toprule
\multicolumn{1}{c|}{Model}  &   \#Param & Validation    & Test  \\
\midrule
AWD-LSTM w/o finetune   & 33M   & 69.1  & 66.0  \\
AWD-LSTM                & 33M   & 68.6  & 65.8  \\
AWD-LSTM $^\dagger$      & 33M   & 46.4  & 44.3  \\
\midrule
MoS w/o finetune        & 35M   & 66.01 & 63.33  \\
MoS                     & 35M   & 63.88 & 61.45  \\
MoS $^\dagger$       & 35M   & \bf42.41  & \bf40.68 \\
\midrule
MoS$+$ w/o finetune & 35M   & 65.33 & 62.66  \\
MoS$+$              & 35M   & 63.71 & 61.28  \\ 
MoS$+$ $^\dagger$   & 35M   & 42.73 & 40.74  \\
\midrule
CT-MoS w/o finetune & 45M   & \bf65.25  & \bf62.21 \\
CT-MoS              & 45M   & \bf62.89  & \bf60.13 \\
CT-MoS $^\dagger$    & 45M   & 42.88     & 40.96 \\
\bottomrule
\end{tabular}
\end{center}
\caption{Perfomrance comparison on the WikiText-2 (WT2) dataset. $\dagger$ indicates using dynamic evaluation.}
\label{LM-WT2}
\end{table}

\subsection{Ablation Studies}
\paragraph{Model size.} 
We conduct ablation studies to control the number of parameters between models in order to eliminate the effects of the parameter size on the performance.  
As mentioned before, we have augmented the original MoS model to have the same parameter size with our model. We have tried two different ways of augmentation. One is increasing only the size of word embedding, while the other is increasing the size of each layer proportionally. Since the latter approach performs better, we only report its performance, and denote it as MoS$+$.
Results are illustrated in Table~\ref{LM-PTB} and Table~\ref{LM-WT2}. Here we notice that MoS+ has a similar perplexity compared to MoS, indicating that directly increasing model parameters cannot improve the performance. Similar observation and results are also reported by \citeauthor{yang2018Mos}. This ablation study shows that the improvements brought by CT-MoS are more than the mere growth of parameters.

\paragraph{Context-Awareness of Temperature.}
To further illustrate the effects of the proposed method, we compare it with conventional temperature scaling methods, that is, using a constant temperature. We experiment with different constant values $\{0.5,1,4\}$ on the MoS model. Results in Table~\ref{LM-context-awareness} show that employing contextual temperature provides better performance than the conventional method. This indicates that the proposed method indeed has merits on adjusting the model parameters, by considering the context and providing a dynamic and optimized temperature value for each token.

\begin{table}[!ht]
\begin{center}
\begin{tabular}{l|cc}
\toprule
\multicolumn{1}{c|}{Model}  & Validation & Test  \\
\midrule
MoS ($\tau=0.5$)    &   60.73   &   58.38   \\
MoS ($\tau=1.0$)    &   58.08   &   55.97   \\
MoS ($\tau=4.0$)    &   57.39   &   55.21   \\ 
CT-MoS  &   \bf56.95    &   \bf54.69    \\
\bottomrule
\end{tabular}
\end{center}
\caption{Effects of context-awareness. Experiments are conducted on the Penn Treebank dataset without fine-tuning.}
\label{LM-context-awareness}
\end{table}

\paragraph{Temperature Normalization Methods.}
Apart from using the Softmax layer to normalize the temperature, we have considered some alternative normalization methods: (a) $\lambda^{Tanh(\mu)}$, which is used in \cite{Lin2018sact}, and gives the range of temperature $(\frac{1}{\lambda}, \lambda)$. And (b) Tanh($\mu$) + $\lambda$, which gives the range $(-1+\lambda, 1+\lambda)$. (c) Softmax, which gives the range $(\sigma(\mu) + \alpha)/\beta $, and is used in our all experiments. Here, $\mu$ denotes the logits of the temperature, that is, $\mu=f(x_{1..t-1}; \rvtheta)^T \cdot  W_{\tau_1} \cdot W_{\tau_2}$. Results are illustrated in Table \ref{LM-PTB-Normalize}. All three methods are evaluated on PTB, and in this ablative study the experiments are conducted without either fine-tuning or dynamic evaluation for conciseness. The experimental results show that applying Softmax as normalization methods outperforms other methods.

\begin{table}[!ht]
\begin{center}
\begin{tabular}{l|cccccl}
\toprule
\multicolumn{1}{c|}{Model}  & hyper-parameter   & Validation    & Test  \\
\midrule
$\lambda^{Tanh(\mu)}$   & $\lambda$ = 4 & 65.11 & 62.21 \\ 
Tanh($\mu$) + $\lambda$ & $\lambda$ = 3 & 61.35 & 58.89 \\
$(\sigma(\mu) + \alpha)/\beta$  & $\alpha$ = 1, $\beta$ = 0.5   & \bf56.95  & \bf54.69 \\
\bottomrule
\end{tabular}
\end{center}
\caption{Different methods for temperature normalization on the Penn Treebank dataset.}
\label{LM-PTB-Normalize}
\end{table}

\subsection{Analysis}
\paragraph{Changes of Temperature During Training.} 
Early in the Introduction section, we have illustrated the change of contextual temperature during the training process, in Figure \ref{fig:T-training}. We see that contextual temperature is capable of learning and tuning itself on a per-token basis, as training proceeds. Hence, each token has a flexible temperature schedule that is optimal for the model.

Besides, we observe that tokens with high temperature values are frequently used, such as ``to'', ``in'', and ``the'', indicating that the model is more confident in predicting them, and thus give them high logit values. However, since these common words don't convey much information in general, the proposed method learns to give them high temperature values so as to rein in their influence when we apply a Softmax layer.
In conclusion, our method has the advantage of determining the schedule automatically on a per-token basis, which is difficult to achieve in traditional constant temperature scheduling methods.

\begin{table*}
\small
\centering
\begin{tabular}{lcccc}
\toprule
\multirow{1}{*}{\bf Reference} & \multicolumn{4}{p{14cm}}{the bonds are rated \textcolor{amaranth}{\bf single-a-1} by moody 's investors service inc. and single-a by \textcolor{azure(colorwheel)}{\bf standard} \& poor 's corp. ...}  \\
\midrule
\multirow{1}{*}{\bf CT-MoS}  & \multicolumn{4}{p{14cm}}{the bonds are rated \textcolor{amaranth}{\bf triple-a} by moody 's and service inc. and $<$unk$>$ by \textcolor{azure(colorwheel)}{\bf standard} \& poor 's corp. ...}  \\
\midrule
\multirow{1}{*}{\bf MoS} & \multicolumn{4}{p{14cm}}{the bonds are rated \textcolor{amaranth}{\bf $<$unk$>$} by moody 'and service inc. and $<$unk$>$ by \textcolor{azure(colorwheel)}{\bf s\&p} \& poor 's corp. ...} \\
\midrule
\multirow{1}{*}{\textcolor{amaranth}{\bf CT-MoS top-4}} & \bf triple-a 0.34 & single-a-2 0.2  & single-a-1 0.15 & single-a-3 0.11 \\
\specialrule{0em}{1.5pt}{1.5pt}
\multirow{1}{*}{\textcolor{amaranth}{\bf MoS top-4}} & \bf $<$unk$>$ 0.28 & \bf triple-a 0.27 & single-a-2 0.24 & single-a-1 0.1 \\
\midrule
\multirow{1}{*}{\textcolor{amaranth}{\bf Temperature}} & \multicolumn{2}{p{6.5cm}}{\bf triple-a $\tau=2+8.34 e^{-5}$} & \multicolumn{2}{p{5.5cm}}{\bf $<$unk$>$ $\tau=2+8.60 e^{-5}$} \\
\midrule
\multirow{1}{*}{\textcolor{azure(colorwheel)}{\bf CT-MoS top-4}} & \bf standard 0.53 & \bf s\&p 0.22 & moody 0.17 & dow 0.02\\
\specialrule{0em}{1.5pt}{1.5pt}
\multirow{1}{*}{\textcolor{azure(colorwheel)}{\bf MoS top-4}} & \bf s\&p 0.4 & moody 0.23 & \bf standard 0.19 & $<$unk$>$ 0.03\\
\midrule
\multirow{1}{*}{\textcolor{azure(colorwheel)}{\bf Temperature}} & \multicolumn{2}{p{6.5cm}}{\bf standard $\tau=2+11.2 e^{-5}$} & \multicolumn{2}{p{6.5cm}}{\bf s\&p $\tau=2+11.4 e^{-5}$} \\
\bottomrule
\end{tabular}
\caption{Analysis of model performance on a sample from the PTB dataset. $\tau$ denotes the temperature of a certain token, or say element, in the temperature vector.}
\label{table:sample-ptb}
\end{table*}

\paragraph{Language Uncertainties.}
Another interesting observation of the proposed contextual temperature is the correlation between a word's relative position in a sentence and its temperature. The formulation of contextual temperature makes it capable to change based on not only the property of the token itself, but also the context around that token. Therefore, a same word might have different temperature values, since it may be in different positions and with different context. 

Figure~\ref{fig:T-pos-15} shows the statistics in terms of means and confidence intervals of temperature vectors at different positions in a sentence. 
To eliminate the effects of samples being too long or too short, 
we analyze sentences that are longer than 15 words and shorter than 25 words. For each of these sentences, we divide it into three disjoint segments: the first 5 words, the middle 5 words and the last 5 words. Then, the three segments are reconstructed to form a ``normalized'' sentence. Such a pre-processing ensures positions of a token 
only have relative effects to the analysis, and will not be effected by the sentence length.

In Figure~\ref{fig:T-pos-15}, we can see that the temperature value is high at beginning positions of a sentence. As the position gets further away from the head of the sentence, the temperature value first has a sharp decrease, followed by another decrease at the positions near the end of the sentence. Our intuition on this observation is as the following: the beginning positions are the region where the model has little confidence, since there is limited information in the history. The model captures this fact, and learns to utilize high temperature values to smooth the probabilities. As the history builds up at later positions, the model becomes more confident about the next token and outputs a probability distribution that is more spiky on the plausible tokens. The formation of the spiky distribution is done by having a lower, or say cooler, temperature. 

A final observation on this analysis is that the confidence intervals of temperature vectors are wider at the beginning but become narrower at the end generally.

\paragraph{Case Studies: Effectiveness.}
We present case studies on the PTB dataset to visualize the effect of contextual temperature. Table~\ref{table:sample-ptb} shows one of the samples, illustrating the differences between the Reference, outputs by CT-MoS, and outputs by MoS. Other samples can be found in Appendix.

We highlight two differences in red and blue colors, respectively. At the location highlighted in red, we see that both CT-MoS and MoS fail to predict the correct answer ``single-a-1'', which refers to a rating for securities. Instead, the CT-MoS model predicts ``triple-a'', referring to the highest rating for securities. Though it is not the same as the reference, it is much closer to the answer. 
MoS, on the other hand, predicts $<$unk$>$, which deviates too much from the ground truth. Taking a closer look at the temperature, the word ``triple-a'' has a temperature of $2+8.34 e^{-5}$, which is a bit smaller than that of the word $<$unk$>$, whose temperature is $2+8.6 e^{-5}$. This contributes to the result that the model chooses ``triple-a'' over  $<$unk$>$. 

Another highlight in this sample, using blue color, is the prediction of the word ``standard''. In this case, the temperature of ``standard'' is smaller than that of ``s\&p'', making the CT-MoS model more likely to predict the word ``standard''.

\paragraph{Case Study: Temperature and Token Position.}
Another aspect to examine how contextual temperature works is to look at the change of the temperature from a specific token across different positions in a sentence. In Table~\ref{table:sample-ptb-same-word}, we take a sample from the PTB dataset, and highlight the occurrences of the word ``mortgage'' in red. We examine its temperature values at each occurrence. 
As the position changes, the proposed method chooses a different temperature value, 
adjusting its confidence of the model's belief. As we have analyzed in Figure~\ref{fig:T-pos-15}, a general trend is that words appearing early in a sentence get larger temperature values, while those appearing near the end get smaller temperature values. In this case, the temperature at the third occurrence of ``mortgage'' is $2+20.1 e^{-5}$, and gradually decreases at subsequent occurrences. Such decrease indicates that the model gains more confidence in making the prediction, most likely due to richer information from the longer history context. 

\begin{table}[!ht]
\small
\begin{center}
\begin{tabular}{lcccc}
\toprule
\multirow{13}{*}{\bf CT-MoS}  & \multicolumn{4}{p{5.1cm}}{loan  \textcolor{amaranth}{\bf mortgage(1)} corp freddie mac posted posted yields on 30-year  \textcolor{amaranth}{\bf mortgage(2)} commitments for delivery within N days $<$eos$>$ N N standard conventional fixed-rate  \textcolor{amaranth}{\bf mortgages(3)} N N N rate rate capped one-year adjustable rate  \textcolor{amaranth}{\bf mortgages(4)} $<$eos$>$ source telerate systems inc $<$eos$>$ federal national  \textcolor{amaranth}{\bf mortgage(5)} association fannie mae posted posted yields on N year  \textcolor{amaranth}{\bf mortgage(6)} commitments for delivery within N days priced at par N N N standard conventional fixed-rate  \textcolor{amaranth}{\bf mortgages(7)} N ...}  \\
\midrule
\multirow{4}{*}{\textcolor{amaranth}{\bf Temperature $\tau$}} & (1) $2+18.9 e^{-5}$ & (2) $2+19.3 e^{-5}$  \\ & (3) $2+20.1 e^{-5}$  & (4) $2+19.2 e^{-5}$ \\ & (5) $2+18.9 e^{-5}$ & (6) $2+19.2 e^{-5}$ \\ & (7) $2+18.2 e^{-5}$ & \\
\bottomrule
\end{tabular}
\end{center}
\caption{Analysis of the temperature value for one specific word but at different positions in a sentence.}
\label{table:sample-ptb-same-word}
\end{table}

\section{Conclusion and Future Work}
\label{sec:Conclusion}
In this paper, we have proposed the \textit{contextual temperature} for training language models. The proposed model is parameterized using a deep neural network, and learns an optimal temperature for each individual class, i.e., each token in vocabulary, based on the history of the context of each token.
Experiments on language modeling datasets show that the proposed models achieve significantly better performances than state-of-the-art models. 
Additionally, our model is capable of generating text that has a higher semantic representation.
In the future, our work opens up new research directions along the line of fully automated temperature mechanism to explore the implementation of context-aware temperature in various NLP tasks, such as summarization, machine translation, and dialogue generation.

\section{Appendix}
\label{sec:Appendix}
\subsection{Partial Derivatives of Loss to Logits}
\label{appendix:grad}
Take the case of two classes as example, assume that the ground-truth class is $i=0$.  In this case, the loss $L$ is $-\ln p_0$, where $p_0$ is the probability of class $0$ output by the model.  The probabilities of the two classes are $\rvp = \sigma(\rvz \oslash \rvtau) = [p_0, p_1]$.  Let $\rvu = \rvz \oslash \rvtau = [u_0, u_1]$, and $u_i = z_i / \tau_i$.  Then, we have
\begin{equation}
    p_i = \frac{e^{u_i}}{e^{u_0} + e^{u_1}}.
\end{equation}

$\rvtau$ is defined in Equation~\ref{eq:def_Tnew}, and is essentially $\sigma(\rvz_\tau)$. So, 
\begin{equation}
    \tau_i = \frac{e^{z_{\tau_i}}}{e^{z_{\tau_0}} + e^{z_{\tau_1}}}.
\end{equation}

The gradients of the loss with respect to logits $z_0$ and $z_1$ are 
\begin{equation}
    \frac{\partial L}{\partial z_i}=\frac{\partial (-\ln p_0)}{\partial z_i}=\left\{
        \begin{array}{lc}
            (p_i-1) \frac{1}{\tau_i} & i = 0 \\
            p_i \frac{1}{\tau_i} & i \neq 0
        \end{array}\right.
\end{equation}

The gradient of $z_{\tau_0}$ is calculated as below
\begin{equation}
\begin{aligned}
    \frac{\partial L}{\partial z_{\tau_0}} 
    &= \frac{\partial L}{\partial p_0} \frac{\partial p_0}{\partial u_0} \frac{\partial u_0}{\partial \tau_0} \frac{\partial \tau_0}{\partial z_{\tau_0}} + 
    \frac{\partial L}{\partial p_0} \frac{\partial p_0}{\partial u_1} \frac{\partial u_1}{\partial \tau_1} \frac{\partial \tau_1}{\partial z_{\tau_0}} \\
    &= -p_1 \frac{-z_0}{\tau_0^2} \frac{e^{z_{\tau_0} + z_{\tau_1}}}{(e^{z_{\tau_0}}+e^{z_{\tau_1}})^2} + p_1 \frac{-z_1}{\tau_1^2} \frac{-e^{z_{\tau_0} + z_{\tau_1}}}{(e^{z_{\tau_0}}+e^{z_{\tau_1}})^2} \\
    &= \frac{1}{\tau_0} p_1 z_0 \tau_1 + \frac{1}{\tau_1} p_1 z_1 \tau_0
\end{aligned}
\end{equation}

Similarly, the gradient of $z_{\tau_1}$ is calculated as
\begin{equation}
\begin{aligned}
    \frac{\partial L}{\partial z_{\tau_1}} 
    &= \frac{\partial L}{\partial p_0} \frac{\partial p_0}{\partial u_0} \frac{\partial u_0}{\partial \tau_0} \frac{\partial \tau_0}{\partial z_{\tau_1}} + 
    \frac{\partial L}{\partial p_0} \frac{\partial p_1}{\partial u_1} \frac{\partial u_1}{\partial \tau_1} \frac{\partial \tau_1}{\partial z_{\tau_1}} \\
    &= -p_1 \frac{-z_0}{\tau_0^2} \frac{-e^{z_{\tau_0} + z_{\tau_1}}}{(e^{z_{\tau_0}}+e^{z_{\tau_1}})^2} + p_1 \frac{-z_1}{\tau_1^2} \frac{e^{z_{\tau_0} + z_{\tau_1}}}{(e^{z_{\tau_0}}+e^{z_{\tau_1}})^2} \\
    &= -\frac{1}{\tau_0} p_1 z_0 \tau_1 - \frac{1}{\tau_1} p_1 z_1 \tau_0
\end{aligned}
\end{equation}

\subsection{Case Studies: Effectiveness}
\label{appendix:effectiveness-samples}
In Table~\ref{table:sample-ptb-more}, we provide with more examples on the PTB dataset. The difference between CT-MoS and MoS are highlighted in red. In the first sample, we see from the red positions that, by appropriately learning the temperature values, the proposed CT-MoS model correctly predicts the token ``results'', while the MoS model predicts ``treasury'', which differs from the ground truth. 

In the second example, we illustrate outputs of a specific token for both the CT-MoS and MoS model. As shown in the example, CT-MOS predicts the same token ``loss'' as the reference sentence. Temperature analysis shows that ``loss'' has a slightly smaller temperature than ``\$'', scaling its probability to much larger. 

We can see similar patterns in the third example. CT-MOS predicts the same token ``nine'' as the reference sentence, while MoS predicts ``first''. Temperature analysis shows that ``nine'' has a slightly smaller temperature than ``first'', scaling its probability to much larger. This helps CT-MoS achieve a better performance in this example.

\begin{table}[!ht]
\small
\begin{center}
\begin{tabular}{lcc} 
\toprule
\multirow{4}{*}{\bf Reference}  & \multicolumn{2}{p{5.5cm}}{these rate indications are n't directly comparable lending practices vary widely by location $<$eos$>$ treasury bills $<$eos$>$ \textcolor{amaranth}{\bf results} of the tuesday ...}  \\
\midrule
\multirow{4}{*}{\bf CT-MoS} & \multicolumn{2}{p{5.5cm}}{these rate indications are n't directly comparable lending practices vary widely by location $<$eos$>$ treasury bills results \textcolor{amaranth}{\bf results} of the monday ...}  \\
\midrule
\multirow{4}{*}{\bf MoS} & \multicolumn{2}{p{5.5cm}}{these rate indications are n't directly comparable lending practices vary widely by location $<$eos$>$ treasury bills results \textcolor{amaranth}{\bf treasury} of the monday ...} \\
\midrule
\textcolor{amaranth}{\bf CT-MoS top-2} & \bf results 0.81 & \bf treasury 0.07 \\ 
\specialrule{0em}{1.5pt}{1.5pt}
\textcolor{amaranth}{\bf MoS top-2} & \bf treasury 0.64 & \bf results 0.09 \\ 
\midrule
\textcolor{amaranth}{$\tau (1e^{-6} + 2)$} & \bf results 8.11 & \bf treasury 8.58 \\
\bottomrule
\specialrule{0em}{3pt}{3pt}
\toprule
\multirow{3}{*}{\bf Reference}  & \multicolumn{2}{p{5.5cm}}{a share compared with a net loss of \$ N million last year after a \textcolor{amaranth}{\bf loss} from discontinued operations ...}  \\
\midrule
\multirow{3}{*}{\bf CT-MoS} & \multicolumn{2}{p{5.5cm}}{a share $<$eos$>$ with \$ loss loss of \$ N million or year $<$eos$>$ the \textcolor{amaranth}{\bf loss} of the operations ...}  \\
\midrule
\multirow{2}{*}{\bf MoS}    & \multicolumn{2}{p{5.5cm}}{a share $<$eos$>$ with \$ \$ loss of \$ N million or year $<$eos$>$ the \textcolor{amaranth}{\bf \$} of the operations ...}  \\
\midrule
\textcolor{amaranth}{\bf CT-MoS top-2} & \bf loss 0.24 & \bf \$ 0.20 \\ 
\specialrule{0em}{1.5pt}{1.5pt}
\textcolor{amaranth}{\bf MoS top-2} & \bf \$ 0.11 & \bf loss 0.09 \\ 
\midrule
\textcolor{amaranth}{$\tau (1e^{-4} + 2)$} & \bf loss 1.40 & \bf \$ 1.62 \\ 
\bottomrule
\specialrule{0em}{3pt}{3pt}
\toprule
\bf Reference  & \multicolumn{2}{p{5.5cm}}{in the \textcolor{amaranth}{\bf nine} months $<$unk$>$ 's net rose N N to \$ N million ...}  \\
\midrule
\bf CT-MoS  & \multicolumn{2}{p{5.5cm}}{the the \textcolor{amaranth}{\bf nine} months the said net income N N to \$ N million ...}  \\
\midrule
\bf MoS & \multicolumn{2}{p{5.5cm}}{the the \textcolor{amaranth}{\bf first} months the said net income N N to \$ N million ...}  \\
\midrule
\textcolor{amaranth}{\bf CT-MoS top-2} & \bf nine 0.17 &  third 0.10 \\ 
\specialrule{0em}{1.5pt}{1.5pt}
\textcolor{amaranth}{\bf MoS top-2} & \bf first 0.14 &  \bf nine 0.12 \\ 
\midrule
\textcolor{amaranth}{$\tau (1e^{-5} + 2)$} & \bf nine 7.37 & \bf first 7.94 \\
\bottomrule
\end{tabular}
\end{center}
\caption{More samples from PTB.}
\label{table:sample-ptb-more}
\end{table}


\begin{thebibliography}{25}
\providecommand{\natexlab}[1]{#1}
\providecommand{\url}[1]{\texttt{#1}}
\providecommand{\urlprefix}{URL }
\expandafter\ifx\csname urlstyle\endcsname\relax
  \providecommand{\doi}[1]{doi:\discretionary{}{}{}#1}\else
  \providecommand{\doi}{doi:\discretionary{}{}{}\begingroup
  \urlstyle{rm}\Url}\fi

\bibitem[{Bahdanau, Cho, and Bengio(2014)}]{bahdanau2014neural}
Bahdanau, D.; Cho, K.; and Bengio, Y. 2014.
\newblock Neural machine translation by jointly learning to align and
  translate.
\newblock \emph{arXiv preprint arXiv:1409.0473} .

\bibitem[{Bengio et~al.(2003)Bengio, Ducharme, Vincent, and
  Jauvin}]{bengio2003neural}
Bengio, Y.; Ducharme, R.; Vincent, P.; and Jauvin, C. 2003.
\newblock A neural probabilistic language model.
\newblock \emph{Journal of machine learning research} 3(Feb): 1137--1155.

\bibitem[{Caccia et~al.(2018)Caccia, Caccia, Fedus, Larochelle, Pineau, and
  Charlin}]{Caccia2018gan}
Caccia, M.; Caccia, L.; Fedus, W.; Larochelle, H.; Pineau, J.; and Charlin, L.
  2018.
\newblock Language GANs Falling Short.
\newblock In \emph{NIPS Workshop}.

\bibitem[{Chen et~al.(2019)Chen, Xie, Huang, Rong, Ding, Huang, Xu, and
  Huang}]{chen2018da}
Chen, C.; Xie, W.; Huang, W.; Rong, Y.; Ding, X.; Huang, Y.; Xu, T.; and Huang,
  J. 2019.
\newblock Progressive Feature Alignment for Unsupervised Domain Adaptation.
\newblock In \emph{CVPR}.

\bibitem[{Dai et~al.(2019)Dai, Yang, Yang, Carbonell, Le, and
  Salakhutdinov}]{Dai2019TransXL}
Dai, Z.; Yang, Z.; Yang, Y.; Carbonell, J.; Le, Q.~V.; and Salakhutdinov, R.
  2019.
\newblock Transformer-XL: Attentive Language Models Beyond a Fixed-Length
  Context.
\newblock In \emph{ACL}.

\bibitem[{Dauphin et~al.(2017)Dauphin, Fan, Auli, and
  Grangier}]{dauphin2017gatedcnn}
Dauphin, Y.~N.; Fan, A.; Auli, M.; and Grangier, D. 2017.
\newblock Language Modeling with Gated Convolutional Networks.
\newblock In \emph{Proceedings of the 34th International Conference on Machine
  Learning-Volume 70}, 933--941. JMLR.

\bibitem[{Guo et~al.(2017)Guo, Pleiss, Sun, and Weinberger}]{Guo2017Cali}
Guo, C.; Pleiss, G.; Sun, Y.; and Weinberger, K.~Q. 2017.
\newblock On Calibration of Modern Neural Netowrks.
\newblock In \emph{ICML}.

\bibitem[{Hinton, Vinyals, and Dean(2014)}]{Hinton2014KD}
Hinton, G.; Vinyals, O.; and Dean, J. 2014.
\newblock Distilling the Knowledge in a Neural Network.
\newblock In \emph{NIPS}.

\bibitem[{Hochreiter and Schmidhuber(1997)}]{hochreiter1997lstm}
Hochreiter, S.; and Schmidhuber, J. 1997.
\newblock Long short-term memory.
\newblock \emph{Neural computation} 9(8): 1735--1780.

\bibitem[{Hu et~al.(2017)Hu, Yang, Liang, Salakhutdinov, and
  Xing}]{Hu2017control}
Hu, Z.; Yang, Z.; Liang, X.; Salakhutdinov, R.; and Xing, E.~P. 2017.
\newblock Toward controlled generation of text.
\newblock In \emph{ICML}.

\bibitem[{Jang, Gu, and Poole(2017)}]{Jang2017gumbel}
Jang, E.; Gu, S.; and Poole, B. 2017.
\newblock Categorical reparameterization with gumbel-softmax.
\newblock In \emph{ICLR}.

\bibitem[{Krause et~al.(2017)Krause, Kahembwe, Murray, and
  Renals}]{Krause2017Dynamic}
Krause, B.; Kahembwe, E.; Murray, I.; and Renals, S. 2017.
\newblock Dynamic evaluation of neural sequence models.
\newblock \emph{arXiv preprint arXiv:1709.07432} .

\bibitem[{Krizhevsky, Sutskever, and Hinton(2012)}]{krizhevsky2012imagenet}
Krizhevsky, A.; Sutskever, I.; and Hinton, G.~E. 2012.
\newblock Imagenet classification with deep convolutional neural networks.
\newblock In \emph{Advances in neural information processing systems},
  1097--1105.

\bibitem[{Lin et~al.(2018)Lin, Sun, Ren, Li, and Su}]{Lin2018sact}
Lin, J.; Sun, X.; Ren, X.; Li, M.; and Su, Q. 2018.
\newblock Learning When to Concentrate or Divert Attention: Self-Adaptive
  Attention Temperature for Neural Machine Translation.
\newblock In \emph{EMNLP}.

\bibitem[{Ma et~al.(2017)Ma, Yin, Liu, Neubig, and Hovy}]{ma2017sqdml}
Ma, X.; Yin, P.; Liu, J.; Neubig, G.; and Hovy, E. 2017.
\newblock Softmax Q-Distribution Estimation for Structured Prediction: A
  Theoretical Interpretation for RAML.
\newblock \emph{arXiv preprint arXiv:1705.07136} .

\bibitem[{Marcus et~al.(1993)Marcus, Marcinkiewicz, , and
  Santorini}]{Marcus1993PTB}
Marcus, M.~P.; Marcinkiewicz, M.~A.; ; and Santorini, B. 1993.
\newblock Building a large annotated corpus of english: The penn treebank.
\newblock In \emph{Computational linguistics}.

\bibitem[{Merity, Keskar, and Socher(2018)}]{Merity2018AWD}
Merity, S.; Keskar, N.~S.; and Socher, R. 2018.
\newblock Regularizing and Optimizing LSTM Language Models.
\newblock In \emph{ICLR}.

\bibitem[{Merity et~al.(2017)Merity, Xiong, Bradbury, and
  Socher}]{Merity2017WT2}
Merity, S.; Xiong, C.; Bradbury, J.; and Socher, R. 2017.
\newblock Pointer Sentinel Mixture Models.
\newblock In \emph{ICLR}.

\bibitem[{Mikolov et~al.(2011)Mikolov, Deoras, Kombrink, Burget, and
  {\v{C}}ernock{\`y}}]{mikolov2011lmtech}
Mikolov, T.; Deoras, A.; Kombrink, S.; Burget, L.; and {\v{C}}ernock{\`y}, J.
  2011.
\newblock Empirical evaluation and combination of advanced language modeling
  techniques.
\newblock In \emph{Twelfth annual conference of the international speech
  communication association}.

\bibitem[{Mikolov et~al.(2010)Mikolov, Karafi{\'a}t, Burget,
  {\v{C}}ernock{\`y}, and Khudanpur}]{mikolov2010rnn}
Mikolov, T.; Karafi{\'a}t, M.; Burget, L.; {\v{C}}ernock{\`y}, J.; and
  Khudanpur, S. 2010.
\newblock Recurrent neural network based language model.
\newblock In \emph{Eleventh annual conference of the international speech
  communication association}.

\bibitem[{Norouzi et~al.(2016)Norouzi, Bengio, Chen, Jaitly, Schuster, Wu, and
  Schuurmans}]{Norouzi2016raml}
Norouzi, M.; Bengio, S.; Chen, Z.; Jaitly, N.; Schuster, M.; Wu, Y.; and
  Schuurmans, D. 2016.
\newblock Reward augmented maximum likelihood for neural structured prediction.
\newblock In \emph{NIPS}.

\bibitem[{Paszke et~al.(2017)Paszke, Gross, Chintala, Chanan, Yang, DeVito,
  Lin, Desmaison, Antiga, and Lerer}]{paszke2017automatic}
Paszke, A.; Gross, S.; Chintala, S.; Chanan, G.; Yang, E.; DeVito, Z.; Lin, Z.;
  Desmaison, A.; Antiga, L.; and Lerer, A. 2017.
\newblock Automatic Differentiation in {PyTorch}.
\newblock In \emph{NIPS Autodiff Workshop}.

\bibitem[{Yang et~al.(2018)Yang, Dai, Salakhutdinov, and Cohen}]{yang2018Mos}
Yang, Z.; Dai, Z.; Salakhutdinov, R.; and Cohen, W.~W. 2018.
\newblock Breaking the Softmax Bottleneck: A High-Rank {RNN} Language Model.
\newblock In \emph{ICLR}.

\bibitem[{Zhang et~al.(2018)Zhang, Yu, Karaman, Zhang, and
  Chang}]{zhang2018heated}
Zhang, X.; Yu, F.~X.; Karaman, S.; Zhang, W.; and Chang, S.-F. 2018.
\newblock Heated-up softmax embedding.
\newblock \emph{arXiv preprint arXiv:1809.04157} .

\bibitem[{Zilly et~al.(2017)Zilly, Srivastava, Koutn{\'\i}k, and
  Schmidhuber}]{zilly2016rnnhighway}
Zilly, J.~G.; Srivastava, R.~K.; Koutn{\'\i}k, J.; and Schmidhuber, J. 2017.
\newblock Recurrent Highway Networks.
\newblock In \emph{ICML}.

\end{thebibliography}

\end{document}